\documentclass[runningheads]{llncs}
\usepackage{algorithm}
\usepackage{algpseudocode}
\usepackage{amsfonts}
\usepackage{amsmath}
\usepackage{amssymb}
\usepackage{booktabs}
\usepackage{bm}
\usepackage{caption}
\usepackage{cite}
\usepackage{color}
\usepackage{dsfont}
\usepackage{enumerate}
\usepackage{float}
\usepackage{graphicx}
\usepackage{epsfig}
\usepackage{epstopdf}
\usepackage{hyperref}
\usepackage{lineno}
\usepackage{mathrsfs}
\usepackage{mdwlist}
\usepackage{multirow}
\usepackage{picinpar}
\usepackage{subfigure}
\usepackage{subfloat}
\usepackage{tabularx}
\usepackage{url}
\usepackage{xcolor}
\def\figurename{\small{Fig.}}
\def\tablename{\small{Table}}
\begin{document}

\title{\textbf{ Boundary feature fusion network for tooth image segmentation}}
\author{Dongping Zhang\inst{1}\and
	Zheng Li\inst{1}\thanks{Corresponding author: s22030810012@cjlu.edu.cn} \and
Fangao Zeng\inst{2} \and
Yutong Wei\inst{3}}
\authorrunning{Li et al.}
\institute{College of Information Engineering, China Jiliang University, Hangzhou, China\and
College of Optoelectronics, China Jiliang University, Hangzhou, China\and 
College of Computer and Control Engineering, Qiqihar University, Qiqihar, China
}
\maketitle              
\begin{abstract}
Tooth segmentation is a critical technology in the field of medical image segmentation, with applications ranging from orthodontic treatment to human body identification and dental pathology assessment. Despite the development of numerous tooth image segmentation models by researchers, a common shortcoming is the failure to account for the challenges of blurred tooth boundaries. Dental diagnostics require precise delineation of tooth boundaries. This paper introduces an innovative tooth segmentation network that integrates boundary information to address the issue of indistinct boundaries between teeth and adjacent tissues. This network's core is its boundary feature extraction module, which is designed to extract detailed boundary information from high-level features. Concurrently, the feature cross-fusion module merges detailed boundary and global semantic information in a synergistic way, allowing for stepwise layer transfer of feature information. This method results in precise tooth segmentation. In the most recent STS Data Challenge, our methodology was rigorously tested and received a commendable overall score of 0.91. When compared to other existing approaches, this score demonstrates our method's significant superiority in segmenting tooth boundaries.

\keywords{Tooth segmentation  \and Boundary information  \and Boundary feature extraction \and Feature cross-fusion.}
\end{abstract}

\section{Introduction}

With the continuous advancement in medical technology, the field of medical image processing has garnered increasing interest, especially in the critical domain of oral medical imaging. At the heart of dental imaging, tooth image segmentation plays a pivotal role in applications ranging from disease detection \cite{r1}, gender determination \cite{r2}, to human body identification \cite{r3}. The goal of tooth image segmentation is to precisely identify and isolate areas of interest, thereby providing dentists with a robust foundation for diagnosis. However, the intricate anatomical structure of teeth, encompassing various components like enamel, dentin, pulp, among others, presents a challenge. The indistinct boundaries among these components \cite{r4} significantly complicate the task of image segmentation. Moreover, the oral environment is fraught with numerous interfering elements such as saliva and reflections, further compromising tooth image quality and exacerbating the segmentation challenge.

In recent years, a plethora of image segmentation techniques have been investigated. These include the automatic segmentation of CBCT dental images using the Otsu threshold and boundary tracking methods \cite{r5}, segmentation based on three-dimensional region merging and histogram thresholds \cite{r6}, and tooth segmentation employing least squares SVM and the mean shift algorithm \cite{r7}, among others. While these traditional methods have proven effective, they tend to be subjective and labor-intensive, particularly when processing large image datasets. Consequently, there has been a shift towards deep learning-based approaches, such as the utilization of the enhanced AlexNet network model for tooth segmentation \cite{r8}, the adoption of the U-Net network for tooth image segmentation \cite{r9}, and automatic tooth segmentation using a two-dimensional coupled shape model in conjunction with the U-Net network \cite{r10}.

However, these methodologies often overlook the crucial aspect of integrating detailed tooth boundary information within the network. This paper addresses this oversight and makes the following contributions:
\begin{itemize}
	\item[$\bullet$] We propose a Boundary Feature Fusion Network aimed at achieving precise segmentation of dental panoramic images.
	\item[$\bullet$] We introduce a Boundary Feature Extraction module based on a reverse attention mechanism, specifically tailored to extract nuanced details of tooth boundaries.
	\item[$\bullet$] We design a Feature Cross-Fusion module to amalgamate boundary detail information with high-level semantic information, thereby facilitating the layered synthesis of a more accurate tooth mask.
\end{itemize}

\section{Method}

This paper introduces a novel boundary feature fusion network, BFFNet, to address the challenges posed by the intricate neural tissue surrounding teeth and the resulting fuzzy boundary segmentation issues. BFFNet is mainly composed of a coding network (E1-E5), a boundary feature extraction module and a feature cross-fusion module. Fig.1 depicts the overall framework of our tooth image segmentation model. The sections that follow provide an in-depth examination of both the overall architecture and the model's critical elements.

Faced with the complexity of nerve tissue around teeth and the resulting blurred boundary segmentation problem, this research proposes the BFFNet, an innovative boundary feature fusion network. This network's design core is divided into two parts: the boundary feature extraction module and the feature cross-fusion module. Fig.1 depicts the overall structure of the tooth image segmentation model. The model's overall architecture and key components are then detailed.

\begin{figure}[htbp]
	\centering
	\includegraphics[width=\textwidth]{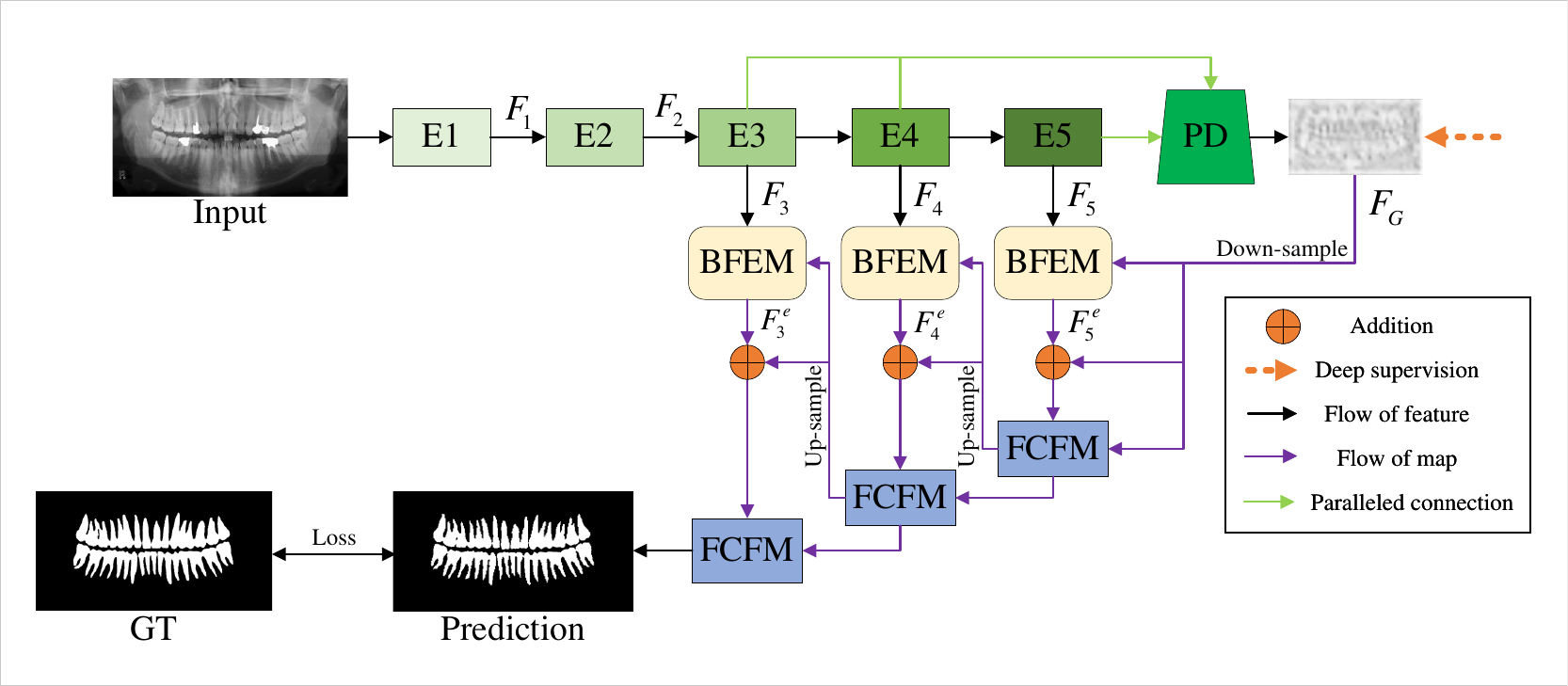}
	\caption{Overview of the proposed BFFNet architecture. It mainly consists of three boundary feature extraction modules and three feature cross-fusion modules.}
	\label{fig:bffnet}
\end{figure}

\subsection{Proposed Method}

In this paper, we propose an advanced segmentation network based on the fusion of tooth boundary features. Specifically, the network takes a tooth image \emph{I} of size h×w as input and uses the backbone network based on the ResNet \cite{r11} architecture to extract five types of features \textit{F\textsubscript{i} }(i=1,...,5) at different levels. These output features\textit{ F\textsubscript{i}} are further subdivided into two categories: low-level features (\textit{F\textsubscript{1}}, \textit{F\textsubscript{2}}) and high-level features (\textit{F\textsubscript{3}}, \textit{F\textsubscript{4}}, and \textit{F\textsubscript{5}}). It is worth noting that we process high-level features through parallel connections and apply the partial decoder (PD) technology proposed in \cite{r12} to obtain the global mapping feature  \textit{F\textsubscript{G}}.

Then, the global mapping features are fed into the first boundary feature extraction module (BFEM) designed by us to extract preliminary boundary information. Subsequently, the extracted boundary information and global mapping features are additively fused to enhance the interaction between details and global information. Thereafter, the fused information is input into the first feature cross-fusion module (FCFM) together with the original global mapping features to achieve deep cross-fusion between multiple features. Subsequent BFEM and FCFM modules follow similar processing flows. In particular, we take the output of the last FCFM as the final tooth mask prediction result.

This novel structural framework enables our network to address the issue of boundary blurring in dental pictures more effectively, providing a more effective and precise tool for oral medical image analysis.

\subsection{Boundary Feature Extraction Module}

In clinical dentistry, dentists must first determine the tooth area before marking the teeth based on information such as position and shape. As mentioned in Section 2.1, the global mapping feature\textit{ \textit{F\textsubscript{G}} }obtained through the output of the last layer of the convolutional neural network mainly contains coarse semantic information of dental tissue but lacks fine local details. To address this issue, our research introduces a boundary feature extraction module based on the reverse attention mechanism \cite{r13}, which aims to extract more refined boundary information from high-level features and effectively transfer this information to promote a more accurate Generation of segmentation masks. Specifically, this module extracts boundary features  by multiplying high-level features \textit{F\textsubscript{i}} (i=3,4,5) and reverse attention weights \textit{W\textsubscript{i}}. Formula 1 describes this process, and the specific expression of the reverse attention weight \textit{W\textsubscript{i}} is shown in formula 2.

\begin{equation}
    F_i^e = \begin{pmatrix} F_i \otimes W_i \end{pmatrix} \oplus F_i
\end{equation}
\begin{equation}
    W_i = E \left.-\left( \boldsymbol{\varepsilon} \left( U_p \left( S_{i+1} \right) \right) \right) \right. 
\end{equation}

\noindent\textit{U\textsubscript{p}}(·) represents the upsampling operation, $\varepsilon$(·) represents the Sigmoid function, E is the all-1 matrix,  $\otimes$ represents the element-wise multiplication operation, and $\oplus$  represents the element-by-element addition operation. Fig.2 shows the details of the entire process.

With this module design, we can improve the detail capture of tooth boundaries while retaining global information, significantly improving the accuracy and detail richness of tooth image segmentation, and providing clinicians with more reliable diagnostic assistance.
\begin{figure}[ht]
\centering
\includegraphics[width=\linewidth]{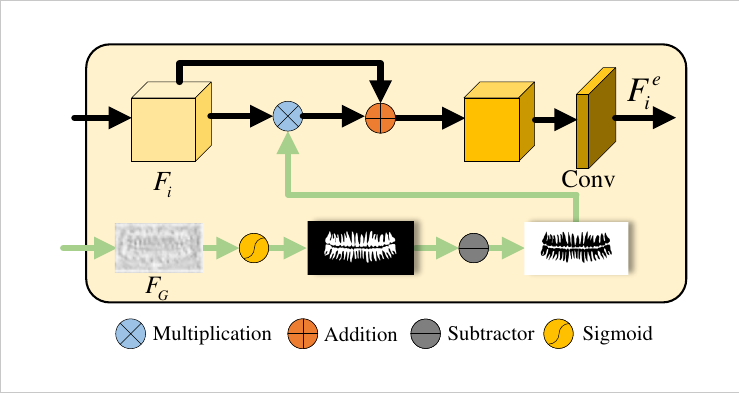}
\caption{Internal structure of BFEM.}
\label{fig:BFEM}
\end{figure}
\subsection{Feature Cross-Fusion Module}

To more efficaciously integrate boundary detail features and realize precise mask prediction, we have designed a feature cross-fusion module (FCFM). The primary aim of this module is to excavate and amalgamate more nuanced semantic information. The architectural details of FCFM are depicted in Fig.3. Within this module, a cross-fusion strategy is employed. The process unfolds as follows: initially, a concatenation operation merges two distinct features, followed by processing the merged features through a branch named 'Local Att.' This step is aimed at deriving features focused on local attention. The computational mechanics of this phase are detailed in formula 3 and  formula 4.

\begin{equation}
    F_{j,j+1}^{local} = \text{Con}\nu_3(F_{j,j+1}) \otimes W_{local} \oplus \text{Con}\nu_3(F_{j,j+1})
\end{equation}
\begin{equation}
    W_{local}=\varepsilon(P-Con\nu_2(\delta(P-Con\nu_1(Con\nu_3(F_{j,j+1})))))
\end{equation}
\begin{equation}
    F_{j,j+1}^u=Con\nu_3(Concat(F_{j,j+1}^{local},Con\nu_3(F_{j,j+1})))
\end{equation}

\noindent\textit{Conv\textsubscript{3}} represents a 3×3 convolution operation, \textit{W\textsubscript{local}} represents the local attention weight, where \textit{P-Conv\textsubscript{i}} represents point-wise convolution, the kernel sizes of P-Conv\textsubscript{1} and P-Conv\textsubscript{2} are  K/t×K×1×1 and K×K/t×1×1 respectively, and T represents the channel reduction rate, K represents the channel size. In addition, $\epsilon$(·) and $\delta$(·) represent the Sigmoid and ReLU activation functions respectively.

Following that, the local attention branch features are spliced with another set of features to finally generate fused features. The process is expressed by formula 5, where Concat represents the concatenation operation.

The FCFM as a whole is intended to improve the model's ability to capture detailed information in tooth images via fine feature processing and fusion, resulting in higher precision results for tooth segmentation and further improving the efficiency and accuracy of tooth image analysis.

\begin{figure}
\centering
\includegraphics[width=\textwidth]{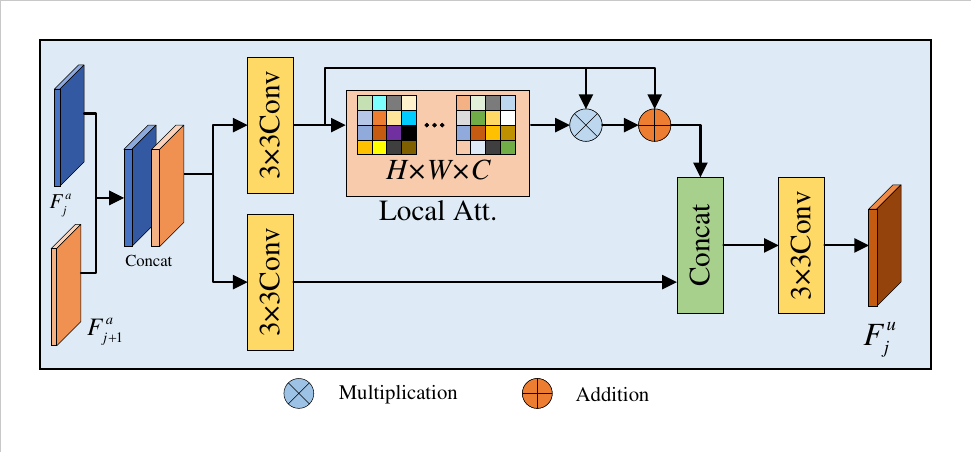}
\caption{Internal structure of FCFM.} \label{fig:FCFMM}
\end{figure}

\subsection{Loss Function}

In this paper, we optimize our boundary feature fusion network (BFFNet) using a specially designed loss function. This loss function is expressed as follows:

\begin{equation}
    Loss=L_{IOU}^w+L_{BCE}^w
\end{equation}

\noindent$L_{IOU}^w$  represents the weighted IOU loss based on global constraints and $L_{BCE}^w$  represents the weighted binary cross entropy (BCE) loss based on local constraints.


These loss function definitions are consistent with those found in the literature \cite{r14,r15}. It is worth noting that we adopt a deep supervision method for the output of three FCFMs (i.e.: $F_{3}^s,F_{4}^s,F_{5}^s$) and the global map $F_{G}$ . The global map obtains the same size as the ground truth map \textit{G} through an upsampling operation. Therefore, the overall loss function of BFFNet can be expressed as:
\begin{equation}
 L_{total} = L\left(G,F_{G}^{up}\right) + L\left(G,F_{3}^{s-up}\right) + L\left(G,F_{4}^{s-up}\right) + L\left(G,F_{5}^{s-up}\right)
\end{equation}
\section{Experiments}
\subsection{Datasets and Comparative Models}
Our experimental study utilizes the STS (Tooth Segmentation Task Based on 2D Panoramic Image) dataset, which was introduced by Zhang et al. \cite{r16} as part of the MICCAI 2023 Challenges. This unique dataset comprises dental panoramic photographs from 106 pediatric patients, ranging in age from 2 to 13 years. It stands as the world's first panoramic photo dataset specifically tailored for pediatric dentistry. The dataset's annotations were meticulously crafted using the efficient interactive segmentation annotation software EISeg and the image annotation software LabelMe. These annotations are primarily geared towards research in caries segmentation and dental disease detection. Additionally, the dataset encompasses dental panoramas from 93 other pediatric patients, as well as 2692 images sourced from three international adult dental datasets. It is important to note that our research utilized only 2000 data entries from the preliminary phase of the STS Data Challenge.

Furthermore, to evaluate the efficacy and superiority of our proposed method, we compared the proposed method with four other leading medical image segmentation methods, including U-Net \cite{r17}, UNet++ \cite{r18}, LDNet \cite{r19}, and CCBANet \cite{r20}. 
\subsection{Evaluation metrics}
To evaluate the model's performance in the experiment, we used the official evaluation indicators of the MICCAI 2023 Challenge, which are the \textit{Dice} coefficient, \textit{IOU}, Hausdorff distance, and a comprehensive score Score. The specific formula is as follows:

\noindent Definition of dice coefficient:
\begin{equation}Dice=\frac{2 * \mid A\cap B\mid}{\mid A\mid+\mid B\mid}\end{equation}
Definition of \textit{IOU}:
\begin{equation}IOU=\frac{(A\cap B)}{(A\cup B)}\end{equation}
A represents predicted mask, and B represents the Ground Truth mask.
The minimum distance between two shapes or curves obtained through the Hausdorff transformation is the two-dimensional Hausdorff distance. It is defined as follows:
\begin{equation}H(d)=\min(\mid x_1-x_2\mid+\mid y_1-y_2\mid)\end{equation}
\noindent$(x_1, y_1)$ and $(x_2, y_2)$  represent the coordinates of two-pixel points, $\mid x_1-x_2\mid$ and $\mid y_1-y_2 \mid$ represent the distance on the corresponding coordinate axis.

The model's performance is primarily determined by the three scoring indicators listed above. The official established a comprehensive score Score to facilitate the observation of the model's performance. Specifically defined as follows:
\begin{equation}Score=w_1*Dice+w_2*IOU+w_3*(1-H(d))\end{equation}
\noindent $w_1$,$ w_2$ and  $w_3$ represent the weight coefficients, which are 0.4, 0.4, and 0.3 respectively.
\subsection{Implementation details}
\textbf{Environment settings.} In this study, we built a specialized development environment, whose detailed configuration is shown in Table 1. We chose the Windows 10 operating system as the primary platform. The core processing unit (CPU) uses Intel(R) Core(TM) i7-8700K, with a clock speed of 3.70GHz. The system memory is 16GB RAM, distributed in two 8GB modules. To handle complex image segmentation tasks, we are equipped with an NVIDIA GeForce RTX 2080 SUPER 8G graphics card. The CUDA version installed in the system is 11.6, which is a key component for deep learning calculations. On the software side, we chose Python 3.7 as the primary programming language. In terms of deep learning frameworks, we used Torch 1.8 and Torch-vision 0.9.

\begin{table}[h]
\centering
\caption{Development environments and requirements.}
\label{tab:my-table}
\begin{tabular}{ll}
\toprule
System                      & Windows 10                            \\
\midrule
CPU                         & Intel(R) Core(TM) i7-8700K CPU @ 3.70GHz \\
\midrule
RAM                         & 16GB                                  \\
\midrule
GPU version                 & NVIDIA GeForce RTX 2080 SUPER 8G      \\
\midrule
CUDA version                & 11.6                                  \\
\midrule
Programming language        & Python 3.7                            \\
\midrule
Deep learning framework     & torch 1.8, torchvision 0.9           \\
\bottomrule
\end{tabular}
\end{table}

\noindent \textbf{Training protocols.} In this study, the training hyperparameters are set as shown in Table 2. The framework is implemented based on PyTorch and trained on an NVIDIA GeForce RTX 2080 SUPER graphics card equipped with 8G video memory.To optimize the overall parameters of the network, we adopted the Adam optimization algorithm and set the learning rate to $10^{-4}$. All input images are uniformly resized to $320\times640$ pixels. To enhance the robustness of the model, we use three different scales $\{0.75, 1, 1.25\}$ to train the model. The entire network adopts an end-to-end training method, and the batch size is set to 4. The entire training process lasts for 300 epochs. In selecting the optimal model, we determine the best model based on the principle of minimum loss. It is important to emphasize that we did not use any unlabeled data during the entire training process to ensure the high quality and consistency of the training set.

\begin{table}[h]
\centering
\caption{Training protocols.}
\label{tab:training-protocols}
\begin{tabular}{ll} 
\toprule
Batch size                  & 4                                    \\
\midrule
Train size                  & \(320 \times 640\)                   \\
\midrule
Total epochs                & 300                                  \\
\midrule
Optimizer                   & Adam                                 \\
\midrule
Initial learning rate (lr)  & 0.0001                               \\
\midrule
Lr decay schedule           & 50                                   \\
\midrule
Training time               & 48 hours                             \\
\midrule
Number of model parameters  & 32.55M                               \\
\bottomrule
\end{tabular}
\end{table}
\subsection{Ablation study}
In this section, we conduct ablation experimental studies on each small component designed to evaluate their specific contribution to the performance of the proposed model. Through these experiments, we aim to reveal the impact of individual components on the overall performance of the model. Table 3 shows the detailed experimental results of the model using various module combinations. Fig.4 provides a thorough box plot analysis, visually representing the impact of different module integrations on model performance.
\newcolumntype{Y}{>{\centering\arraybackslash}X}
\begin{table}[h]
\centering
\caption{Results of ablation experiments on various BFFNet components. The best outcomes are highlighted. ↑ indicates that the higher the value, the better the performance, whereas ↓ indicates that the lower the value, the better the performance.
}
\label{tab:model-comparison}
\begin{tabularx}{\textwidth}{l|Y Y Y Y}
\hline
Settings & Dice\textsuperscript{$\uparrow$} & IOU\textsuperscript{$\uparrow$} & HD\textsuperscript{$\downarrow$} & Score\textsuperscript{$\uparrow$} \\
\hline
Backbone & 0.6868 & 0.8515 & 0.2097 & 0.7673 \\
Backbone+BFEM & 0.7175 & 0.8902 & 0.0395 & 0.8422 \\
Backbone+FCFM & 0.7212 & 0.8944 & 0.0269 & 0.8487 \\
\textbf{Backbone+BFEM+FCFM} & \textbf{0.7911} & \textbf{0.9848} & \textbf{0.0174} & \textbf{0.9061} \\
\hline
\end{tabularx}
\end{table}
\begin{figure}[htbp]
\centering
\includegraphics[width=\linewidth]{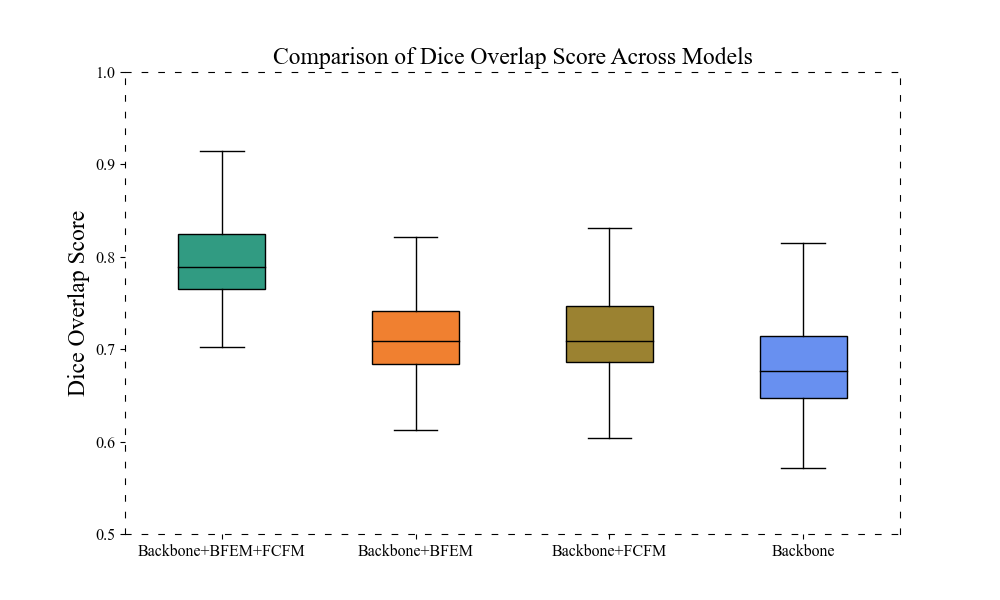}
\caption{Box plot under different module combinations.}
\label{fig:Box plot}
\end{figure}

\noindent\textbf{BFEM} \textbf{of Effectiveness}. We study the importance of the boundary feature extraction module. By comparing the data in the second and third rows in Table 3, we find that the performance of the model is significantly improved after integrating BFEM in the backbone network. Specifically, the \textit{Dice} coefficient increased from 0.6868 to 0.7175, the \textit{IOU} increased from 0.8515 to 0.8902, and the overall score (Score) increased from 0.7673 to 0.8422. These results clearly show that BFEM plays a key role in improving model performance. 

\noindent\textbf{FCFM of Effectiveness}. Further, we studied the impact of FCFM on model performance. By comparing the experimental data in the second and fourth rows in Table 3, we can find that the performance of the model has also been significantly improved after integrating FCFM: the \textit{Dice} coefficient increased to 0.7212, the \textit{IOU} increased to 0.8944, and the total score increased to 0.8487. These experimental results fully demonstrate the importance of FCFM in improving model performance.

\noindent\textbf{BFEM} \textbf{\&} \textbf{FCFM of Effectiveness}. To verify the effectiveness of the combination of BFEM and FCFM, we studied the performance of the combined model. Compare the experimental data in the fifth row and the second to fourth rows in Table 3. We can find that the performance of the model has been significantly improved after adding the two modules, especially the two evaluation indicators of \textit{Dice} and \textit{IOU} have been improved by 7\% and 9\% respectively, and \textit{HD} has reached the lowest value. Further, the boxplot in Fig.4 reveals that the model incorporating both BFEM and FCFM components yields the highest median \textit{Dice} Overlap Score. This suggests that the synergistic interaction of these components significantly bolsters the model's segmentation performance. Additionally, the model's interquartile range being relatively narrow indicates consistent and stable performance. Consequently, it can be inferred that the model integrating both BFEM and FCFM outperforms all other configurations. 
\section{Results and discussion}
BFFNet is a novel automatic segmentation model for tooth images that we created. The experimental results will be analyzed and discussed further below.
\subsection{Quantitative results}
\newcolumntype{C}{>{\centering\arraybackslash}X}

In evaluating and comparing model performance, we consider several key factors in segmentation tasks, including important aspects such as accuracy, consistency, and shape matching. Table 4 shows the quantitative comparison results between our proposed method and the other four methods on four key evaluation indicators.

\begin{table}[h]
	\centering
	\caption{Model Comparison Results.}
	\label{tab:model-comparison}
	\begin{tabularx}{\textwidth}{l|X X X X}
		\hline
		Method &  Dice\textsuperscript{$\uparrow$} & IOU\textsuperscript{$\uparrow$} & HD\textsuperscript{$\downarrow$} & Score\textsuperscript{$\uparrow$} \\
		\hline
		UNet & 0.5832 & 0.7232 & 0.2428 & 0.6774 \\
		UNet++ & 0.5939 & 0.7365 & 0.2429 & 0.6857 \\
		LDNet & 0.6951 & 0.8564 & 0.2202 & 0.7689 \\
		CCBANet & 0.7273 & 0.9023 & 0.0599 & 0.8436 \\
		\hline
		BFFNet(Our) & \textbf{0.7911} & \textbf{0.9848} & \textbf{0.0174} & \textbf{0.9061} \\
		\hline
	\end{tabularx}
\end{table}
\begin{figure}[htbp]
	\centering
	\includegraphics[width=0.7\linewidth]{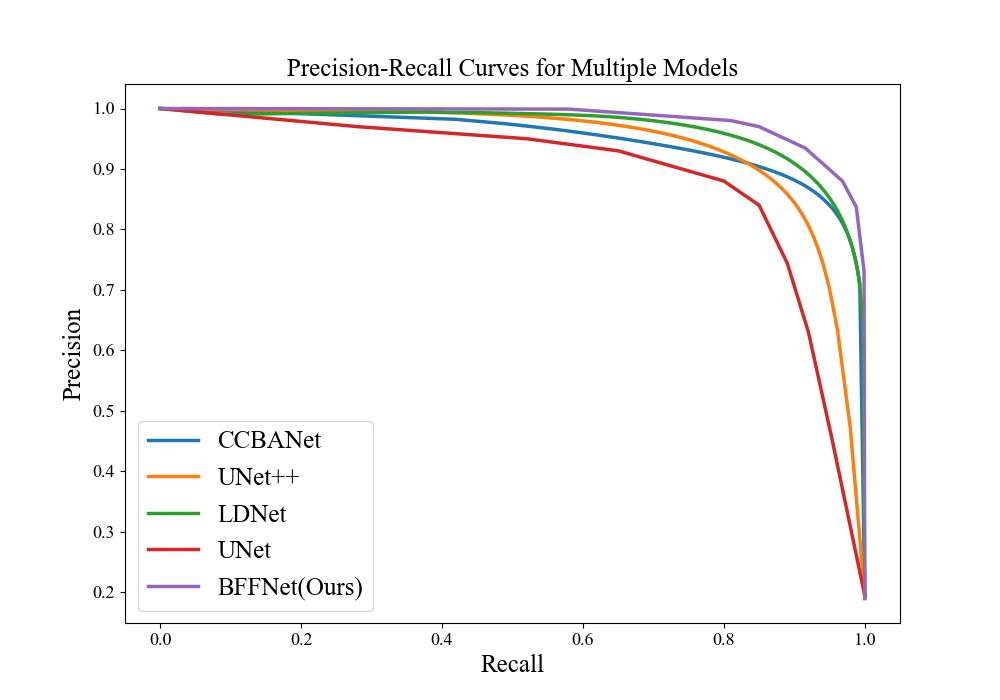}
	\caption{Precision-Recall curves of BFFNet and other four models.}	\label{fig:PR}
\end{figure}
Fig.5 clearly shows that our model achieves significant improvements in segmentation performance due to the use of high-performance components. In addition, the precision-recall curve of BFFNet covers the curves of other segmentation models, which not only highlights its excellent performance but also further verifies the superiority of our proposed tooth segmentation model in various performance indicators.
\subsection{Qualitative results}
The segmentation results of BFFNet and its four models are output and displayed. The detailed comparison results are shown in Fig.6 below.
\begin{figure}[H]
	\rotatebox[origin=lt]{90}{Input}
	\subfigure{
		\begin{minipage}[t]{0.3\textwidth}
			\centering
			\includegraphics[width=1\textwidth]{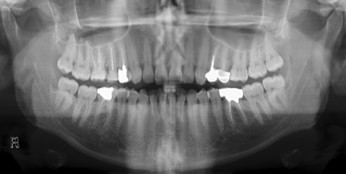}
		\end{minipage}
	}
	\subfigure{
		\begin{minipage}[t]{0.3\textwidth}
			\centering
			\includegraphics[width=1\textwidth]{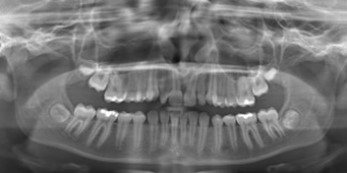}
		\end{minipage}
	}
	\subfigure{
		\begin{minipage}[t]{0.3\textwidth}
			\centering
			\includegraphics[width=1\textwidth]{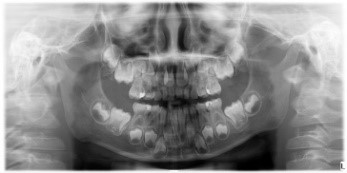}
		\end{minipage}
	}
	
	\rotatebox[origin=lt]{90}{UNet}
	\subfigure{
		\begin{minipage}[t]{0.3\textwidth}
			\centering
			\includegraphics[width=1\textwidth]{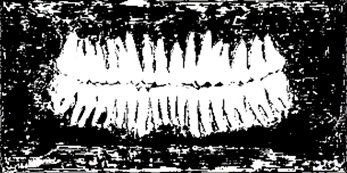}
		\end{minipage}
	}
	\subfigure{
		\begin{minipage}[t]{0.3\textwidth}
			\centering
			\includegraphics[width=1\textwidth]{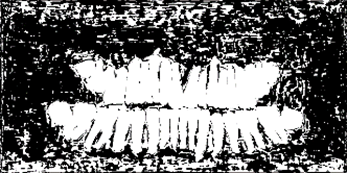}
		\end{minipage}
	}
	\subfigure{
		\begin{minipage}[t]{0.3\textwidth}
			\centering
			\includegraphics[width=1\textwidth]{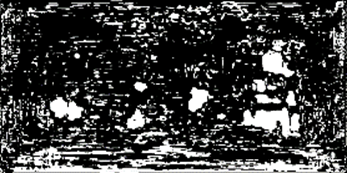}
		\end{minipage}
	}
	
	\rotatebox[origin=lt]{90}{UNet++}
	\subfigure{
		\begin{minipage}[t]{0.3\textwidth}
			\centering
			\includegraphics[width=1\textwidth]{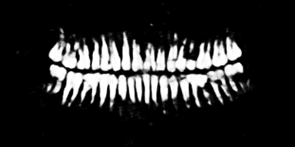}
		\end{minipage}
	}
	\subfigure{
		\begin{minipage}[t]{0.3\textwidth}
			\centering
			\includegraphics[width=1\textwidth]{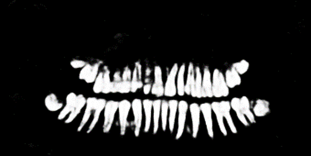}
		\end{minipage}
	}
	\subfigure{
		\begin{minipage}[t]{0.3\textwidth}
			\centering
			\includegraphics[width=1\textwidth]{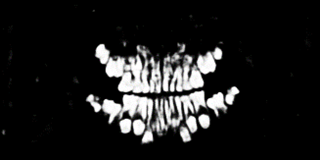}
		\end{minipage}
	}
	
	\rotatebox[origin=lt]{90}{LDNet}
	\subfigure{
		\begin{minipage}[t]{0.3\textwidth}
			\centering
			\includegraphics[width=1\textwidth]{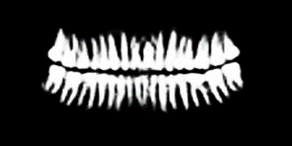}
		\end{minipage}
	}
	\subfigure{
		\begin{minipage}[t]{0.3\textwidth}
			\centering
			\includegraphics[width=1\textwidth]{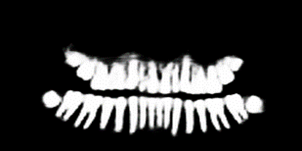}
		\end{minipage}
	}
	\subfigure{
		\begin{minipage}[t]{0.3\textwidth}
			\centering
			\includegraphics[width=1\textwidth]{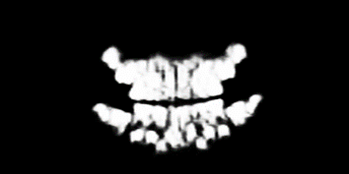}
		\end{minipage}
	}
	
	\rotatebox[origin=lt]{90}{CCBANet}
	\subfigure{
		\begin{minipage}[t]{0.3\textwidth}
			\centering
			\includegraphics[width=1\textwidth]{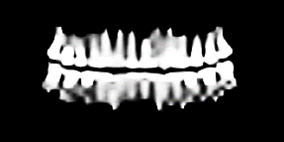}
		\end{minipage}
	}
	\subfigure{
		\begin{minipage}[t]{0.3\textwidth}
			\centering
			\includegraphics[width=1\textwidth]{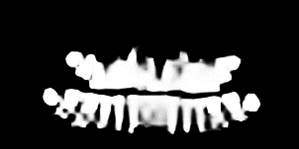}
		\end{minipage}
	}
	\subfigure{
		\begin{minipage}[t]{0.3\textwidth}
			\centering
			\includegraphics[width=1\textwidth]{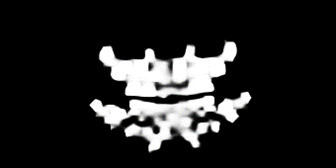}
		\end{minipage}
	}
	
	\rotatebox[origin=lt]{90}{Ours}
	\subfigure{
		\begin{minipage}[t]{0.3\textwidth}
			\centering
			\includegraphics[width=1\textwidth]{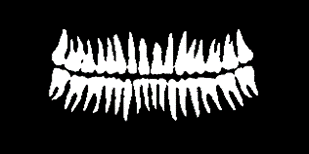}
		\end{minipage}
	}
	\subfigure{
		\begin{minipage}[t]{0.3\textwidth}
			\centering
			\includegraphics[width=1\textwidth]{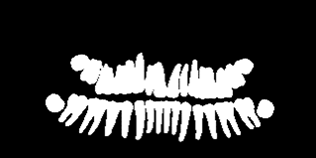}
		\end{minipage}
	}
	\subfigure{
		\begin{minipage}[t]{0.3\textwidth}
			\centering
			\includegraphics[width=1\textwidth]{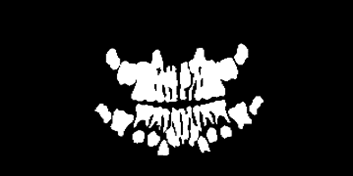}
		\end{minipage}
	}
	
	\rotatebox[origin=lt]{90}{GT}
	\subfigure{
		\begin{minipage}[t]{0.3\textwidth}
			\centering
			\includegraphics[width=1\textwidth]{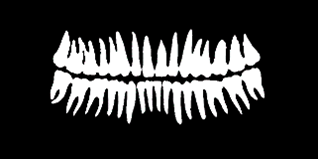}
		\end{minipage}
	}
	\subfigure{
		\begin{minipage}[t]{0.3\textwidth}
			\centering
			\includegraphics[width=1\textwidth]{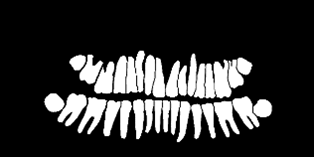}
		\end{minipage}
	}
	\subfigure{
		\begin{minipage}[t]{0.3\textwidth}
			\centering
			\includegraphics[width=1\textwidth]{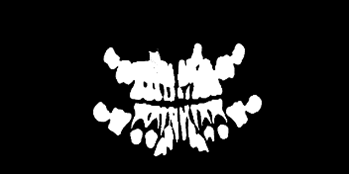}
		\end{minipage}
	}
	
	\centering
	\caption{Qualitative results of different methods.}
	\label{fig:P_end}
\end{figure}
Examining the visualization results reveals that our proposed model has significant advantages in performing boundary segmentation, as well as a robust anti-interference capability against background noise. This level of performance not only highlights the model's ability to discern fine details but also demonstrates its formidable ability to maintain the image's overall stability. Such characteristics highlight the model's effectiveness and dependability in complex image-processing scenarios.

In addition, we find that the four compared models can complete effective segmentation and the BFFNet proposed in this paper achieves the best performance, especially in boundary segmentation, where our model is more advantageous. However, we find that UNet has the worst segmentation results, which is mainly attributed to our limited hardware resources, and through repeated experiments, the results show that the UNet model has never been able to achieve considerable results.

\subsection{Limitation and Future Work}


This paper mainly employs dental images to develop a fully supervised model. However, we did not consider the potential impact of label inaccuracies in the data on model prediction results. In future research, we will consider more about developing tooth image segmentation models based on weak supervision or unsupervised learning.

\section{Conclusion}

In this paper, we address the problem of boundary blur in tooth images and propose an innovative \textbf{B}oundary \textbf{F}eature \textbf{F}usion \textbf{N}etwork (\textbf{\textit{BFFNet}}) that can be used to accurately segment teeth on panoramic X-rays. This method performed admirably in the preliminary competition of MICCAI 2023 Challenges. The experimental results fully demonstrate the proposed method's significant advantages over other medical image segmentation technologies, and it has the potential to become an important tool to assist clinicians in rapid diagnosis. 
Specifically, the tooth images are first processed by the coding network (E1-E5) to obtain features at different scales, and then the global mapping features are output by the partial decoder after processing the advanced features through parallel concatenation. Secondly, the boundary feature extraction module based on the inverse attention mechanism is used to obtain the edge detail information of the teeth. Finally, combined with the feature cross-fertilization module, the effective transmission and fusion of features are realized, so as to achieve the accurate positioning and segmentation of teeth.
Although this study focuses on the detailed segmentation of dental images, the proposed method also has wide applicability in other related fields, such as cell nucleus segmentation \cite{r21}, segmentation of lung infection \cite{r22}, etc., showing its application in medical images. Broad potential and application prospects in processing fields.

\section*{Acknowledgements}
This work was supported by Zhejiang Key R $\&$ D Project of China (2024C01102, 2024C01108, 2023C01030, 2022C01082).

\bibliographystyle{splncs04}
\bibliography{reference}

\end{document}